\DocumentMetadata{
  pdfversion=2.0,pdfstandard=ua-2,
  testphase={phase-III,firstaid,math,title}
}

\documentclass[sigconf,screen]{acmart-tagged}

\AtBeginDocument{%
  }

\setcopyright{cc}
\setcctype{by}
\acmDOI{10.1145/3757279.3788806}
\acmYear{2026}
\copyrightyear{2026}
\acmISBN{979-8-4007-2128-1/2026/03}
\acmConference[HRI '26]{Proceedings of the 21st ACM/IEEE International Conference on Human-Robot Interaction}{March 16--19, 2026}{Edinburgh, Scotland, UK}
\acmBooktitle{Proceedings of the 21st ACM/IEEE International Conference on Human-Robot Interaction (HRI '26), March 16--19, 2026, Edinburgh, Scotland, UK}
\received{2025-10-07}
\received[accepted]{2025-12-09}

\begin{document}

\title{Quest2ROS2: A ROS 2 Framework for Bi-manual VR Teleoperation}

\author{Jialong Li}
\authornote{Both authors contributed equally to this research.}
\orcid{0009-0009-0101-845X}
\affiliation{%
  \institution{Lund University}
  \city{Lund}
  \country{Sweden}
}
\email{jialong.li@cs.lth.se}

\author{Zhenguo Wang}
\authornotemark[1]
\orcid{0009-0001-5831-7954}
\affiliation{%
  \institution{Lund University}
  \city{Lund}
  \country{Sweden}
}
\email{zh6765wa-s@student.lu.se}

\author{Tianci Wang}
\orcid{0009-0001-2825-4120}
\affiliation{%
  \institution{Lund University}
  \city{Lund}
  \country{Sweden}
}
\email{ti4050wa-s@student.lu.se}

\author{Maj Stenmark}
\orcid{0000-0001-5822-876X}
\affiliation{%
  \institution{Lund University}
  \city{Lund}
  \country{Sweden}
}
\email{maj.stenmark@cs.lth.se}

\author{Volker Krueger}
\orcid{0000-0002-8836-8816}
\affiliation{%
  \institution{Lund University}
  \city{Lund}
  \country{Sweden}
}
\email{volker.krueger@cs.lth.se}


\begin{abstract}
Quest2ROS2 is an open-source ROS2 framework for bi-manual teleoperation designed to scale robot data collection. Extending Quest2ROS, it overcomes workspace limitations via relative motion-based control, calculating robot movement from VR controller pose changes to enable intuitive, pose-independent operation. The framework integrates essential usability and safety features, including real-time RViz visualization, streamlined gripper control, and a pause-and-reset function for smooth transitions. We detail a modular architecture that supports "Side-by-Side" and "Mirror" control modes to optimize operator experience across diverse platforms. Code is available at: \url{https://github.com/Taokt/Quest2ROS2}.
\end{abstract}



\maketitle

\section{Introduction}

 \begin{figure}[th]
  \centering
  \includegraphics[width=\linewidth]{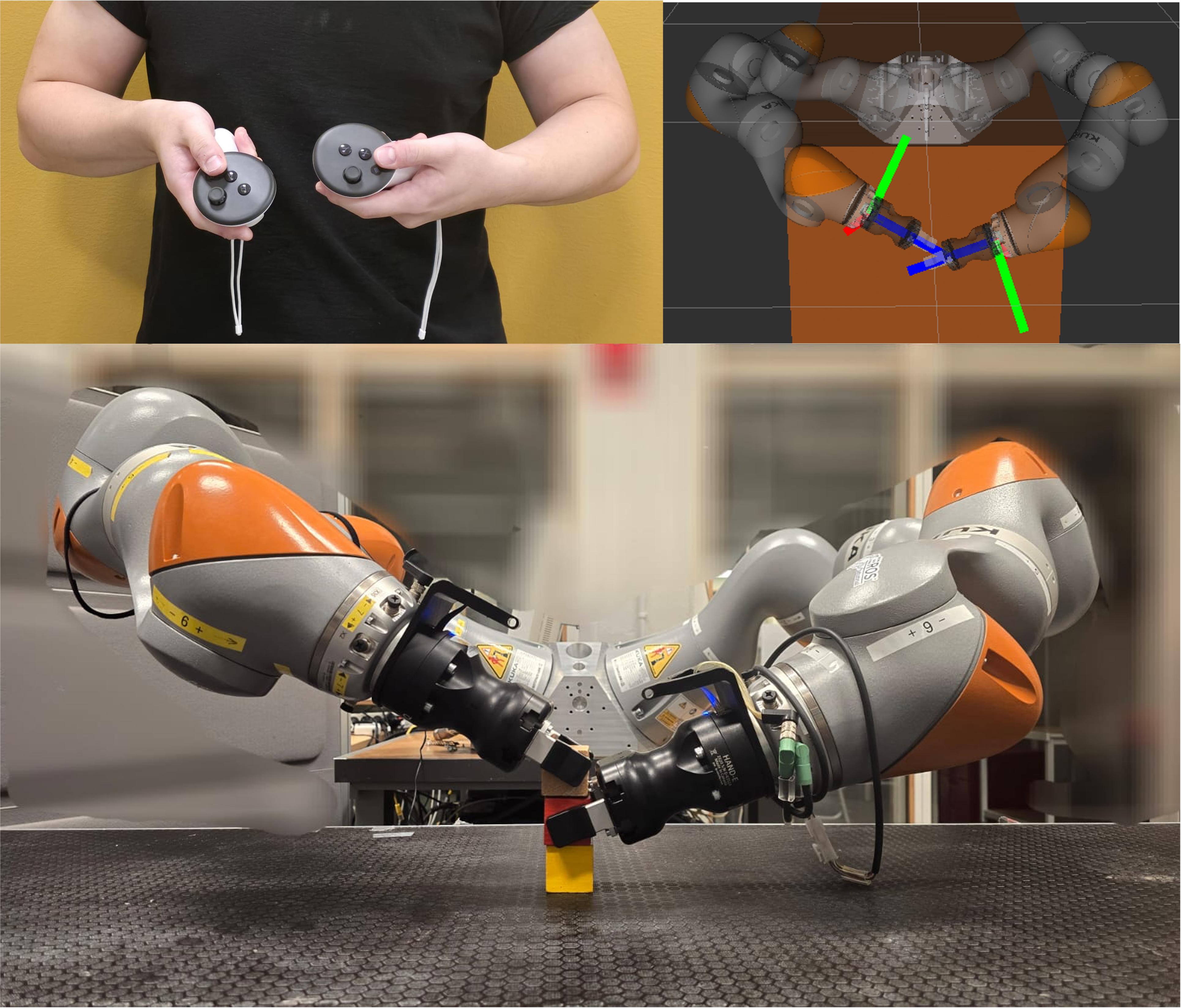}
  \caption{Example implementation of Quest2ROS2 framework on a bi-manual Kuka iiwa7 system. \textit{Top Left}: Potential pose of the operator while teleoperating, the controllers can be grasped in any arbitrary pose. \textit{Top Right}: Visualization of the commands being streamed from the controllers to the robot. \textit{Bottom}: The block-stacking task being performed via teleoperation by the operator using the system.}
  \Description{This is figure 1}
  \label{fig:3in1}
\end{figure}

Robot teleoperation technology has wide-ranging applications in many areas of robotics research, for examples, medical robotics \cite{gentile2023manipulation} and manipulation in hazardous environments  \cite{mehrdad2020review}. With the rise of robot learning methods, represented by vision-language-action models \cite{brohan2022rt, team2024octo, kim2024openvla}, the robotics community has increasingly focused on the collection and quality of robot data \cite{vuong2023open, kim2024survey}, similar to the trends in natural language processing and computer vision. Intuitive methods such as kinesthetic teaching where the human moves the robot arm cannot be used for VLA as data generated in this way will have human in the image. Consequently, robot teleoperation methods specifically for robotic manipulation are receiving increasing attention, with advanced data augmentation techniques like Diffusion models being explored. However their effectiveness remains fundamentally limited by the quality and scale of the initial, reliable teleoperation data \cite{zhang2025training}.

The ongoing effort in this field aims to explore simpler and more user-friendly robot teleoperation methods for data collection. A subset of these efforts focuses on establishing intuitive leader-follower systems like ALOHA \cite{zhao2023learning}, attempting to make robot teleoperation straightforward and easy to operate. However, these systems require a custom setup of corresponding robotic hardware and are difficult to generalize to robots with complex configurations. Another set of works is dedicated to using teach pendants or joy-sticks for this task \cite{shu2022platform, zhang2020augmented}, but due to the difference between the robot's degrees of freedom and configuration and those of a human, these control methods often lack intuitive alignment with the user's expectations. Furthermore, some works adopt a 3D mouse to directly control the movement of the robot's end-effector in Cartesian space \cite{dhat2024using, kim2022telerobotic}. While this method is quite intuitive and cost-effective for operating a single arm, extending it to dual-arm manipulation places certain demands on the operator's coordination skills. 

We attempt to establish a teleoperation system that is potentially usable for dual-arm robots. Based on this goal, we consider VR equipment to be a natural choice for this task due to its ease of operation and low cost. Previous work has explored the feasibility of VR in various robotic teleoperation tasks \cite{luo2024user, crepaz2025virtual, welle2024Quest2ROS}. Among these, Quest2ROS \cite{welle2024Quest2ROS} provides software that enables communication between Meta Quest 2/3 VR devices\footnote{\url{https://en.wikipedia.org/wiki/Meta_Quest_3}} and Robot Operating System(ROS). Its feature of transmitting VR controller data allows it to be applied to the control of robot end-effectors. However, although their work includes basic ROS-side code support, the functionality provided by the original code framework is quite limited and is confined to ROS 1, meaning that users still require considerable upfront development before undertaking a specific task. 

Recently, frameworks like OPENTEACH \cite{OPENTEACH} and BEAVR \cite{BEAVR} have utilized Quest-based hand tracking to enable low-cost teleoperation without the bulky leader-follower hardware required by ALOHA \cite{zhao2023learning}. While similar in goal, our work prioritizes controller-based tracking to ensure superior stability and flexibility. Unlike hand-tracking systems that struggle with pose recognition at certain angles, native controllers offer robust tracking and "relaxed" usage conditions, allowing operators to work without wearing the HMD continuously. Furthermore, we maintain a modular software design by decoupling teleoperation logic from robot-specific IK solvers or remapping layers. By leveraging standard community controller modules instead of integrated solvers, our framework simplifies migration across diverse robotic platforms, offering a more adaptable solution for varied research environments.

\section{Quest2ROS2}
Building upon the Quest2ROS application, this work aims to provide a more functionally complete, easily extensible, and user-friendly open-source code framework on ROS 2 for dual-arm robots, which we name Quest2ROS2. Specifically, the original Quest2ROS architecture primarily consists of software running on the VR device and a middleware layer on the ROS side that processes the controller information fed back by the software to facilitate downstream tasks. Our work primarily focuses on providing a more functionally comprehensive and development-accelerating ROS 2 middleware framework, which now is currently developed based on controlling the end-effector movement in Cartesian space. While retaining the native Quest2ROS functionality, we primarily introduce the following additional key features: control based on the relative motion of the VR controller, end-effector command visualization, and minor functionalities such as gripper state toggle and pose stream state toggle. An example implementation of Quest2ROS2 can be seen at Figure \ref{fig:3in1}.

\subsection{Relative-motion-based End-Effector control}
\begin{figure}[t]
  \centering
  \includegraphics[width=\linewidth]{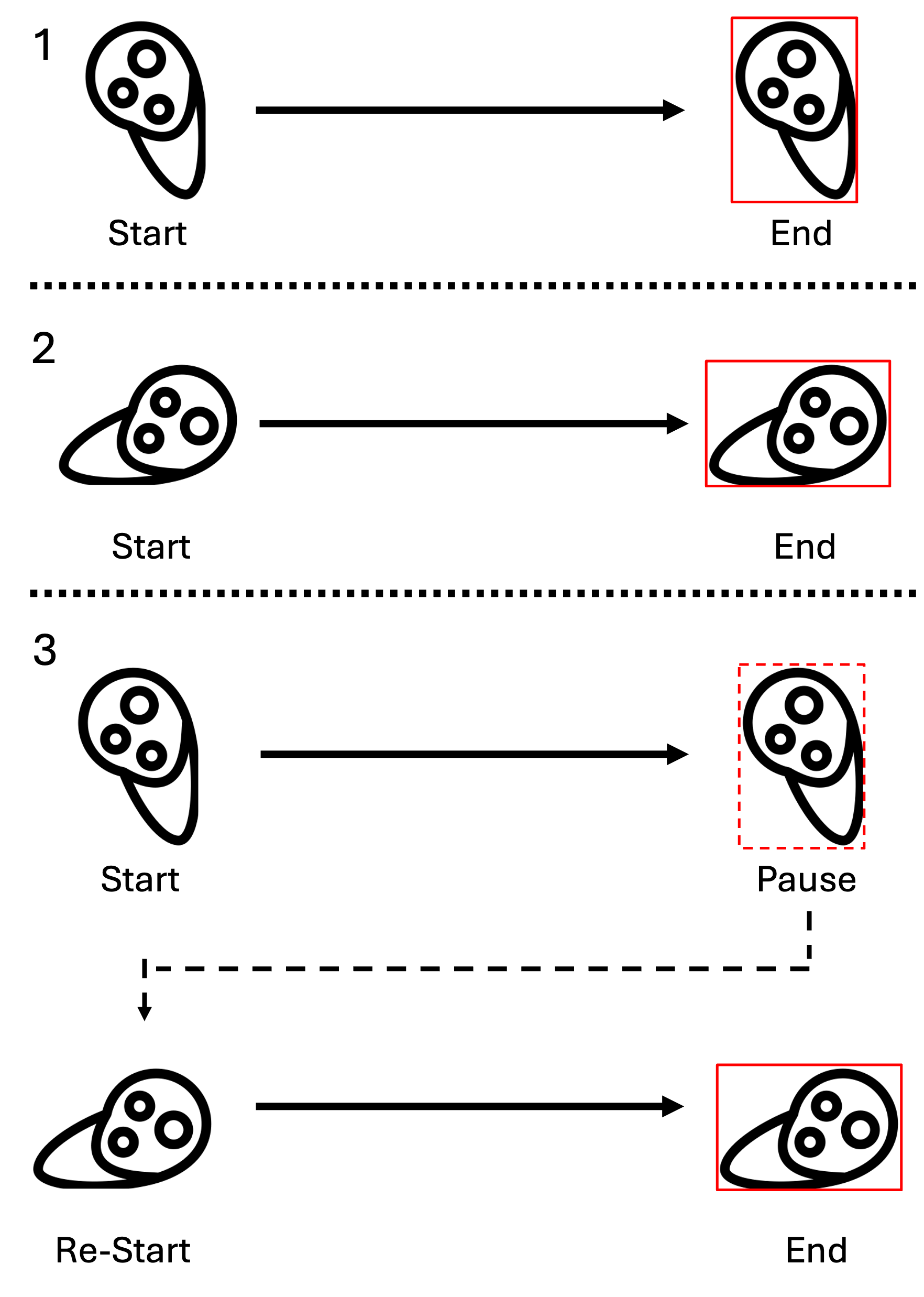}
  \caption{Conceptual demonstration of relative motion based control. Under this control logic, Movement 1 and Movement 2 produce equivalent effects on the End-Effector (EE), causing the EE to translate the same distance in the same direction, and Movement 3 will push the EE to move twice the distance in the same direction.}
  \label{fig:movementmethod}
\end{figure}
Since robot configurations differ from those of humans, real-time, direct control of robot joint movements is often overly challenging for human teleoperation. To circumvent this issue, similar to Quest2ROS, we opt for direct control of the robot's end-effector position with the assistance of other underlying robot controllers in community such as Cartesian impedance controller \footnote{\url{https://github.com/idra-lab/kuka_lbr_control}}.

Specifically, the basic control method provided in the Quest2ROS requires aligning the controller's coordinate frame with the robot's base frame before control begins, and subsequently makes the robot's end-effector directly replicate the controller's pose. While this operation logic is intuitive, we found that in practical applications, it is not sufficient for more flexible robot control and convenient human operation, and it suffers from the following drawbacks:

\textbf{Limited Workspace}: Under the basic Quest2ROS control logic, the robotic arm attempts to replicate the controller's actions, and thus the robot's workspace is inherently constrained by the operator's physical reach. For large-scale, high-reach manipulators such as the Kuka iiwa7, this significantly hinders the robot's ability to complete complex manipulation tasks.

\textbf{Limited Action Feasibility}: Similar to robots, human hand movements are constrained by the degrees of freedom imposed by their configuration. Actions that specifically require high dexterity in the wrist are often difficult for untrained operators to perform, yet such movements are often unavoidable for high-DOF robotic arms. For example, the robot arm can normally rotate around 360 degrees at the wrist, which is impossible for a human to do in one motion.

\textbf{Difficult to adapt to Dual-Arm Tasks}: For single-arm tasks, operators can potentially compensate for workspace limitations or kinematic singularities by adjusting their body posture—such as tilting the torso or moving the lower body. However, such compensatory movements are often insufficient for simultaneous dual-arm operation and may even introduce significant safety risks. Because the two arms share a common operator frame, a postural change intended to assist the motion of one arm inevitably alters the pose and trajectory of the other. This coupling increases the cognitive load on the operator and complicates precise bi-manual coordination.

To resolve this, we propose a control logic where the robotic arm’s movement is determined by the controller’s motion relative to its own startup pose. Instead of replicating absolute coordinates, the end-effector reproduces the controller's relative pose changes, as illustrated in Figure \ref{fig:movementmethod}. We further enhance this logic with a button-triggered "pause-and-resume" functionality to update relative poses dynamically. Consequently, Quest2ROS2 ensures that end-effector control remains entirely independent of the controller's initial physical orientation. This allows operators to select an ergonomic starting position and make seamless adjustments whenever physical limits are reached, all without altering their main body posture or compromising the stability of the dual-arm system

\subsection{Visualization in Simulation}
In addition to directly observing the robot's movement under the influence of the controller, the visualization of the controller's commands is indispensable for facilitating usage and for debugging during the development phase. We introduced visualization markers in Rviz \cite{kam2015rviz} to aid user understanding, as shown at the top right of Figure \ref{fig:3in1}. These markers real-time display the target pose of the robot's end-effector resulting from the commands issued by the controller. To differentiate the markers for the left and right arms, we place the root node of the marker at the root of the respective left and right end-effectors.

\subsection{Button Functionality}
Utilizing the capability of Quest2ROS to also transmit VR controller button information, we have added two functionalities to the upper and lower buttons on the controller.

\textbf{1. Upper Button: Gripper Control.}
We introduced the basic logic for determining gripper opening and closing. It is worth noting that the actual operation of the gripper's open/close command is achieved by controlling the distance between the gripper fingers, which is currently commanded as either the maximum or minimum distance. Therefore there is full potential to introduce more flexible control logic that can adjust the finger position of the gripper flexibly. The current implementation is based on the Robotiq Hand-E gripper with the corresponding driver\footnote{\url{https://github.com/AGH-CEAI/robotiq_hande_driver}}.

\textbf{2. Lower Button: Pause-and-Restart of Pose Command stream.}
As mentioned above, we added the functionality to pause and reactivate command transmission using the downward button to flexibly leverage the start-pose-independent nature of our control method. Furthermore, each time the command transmission is paused and restarted, the function resets the controller's initial pose to its current pose, and instantly aligns the virtual target (Marker) to the robot's actual pose, thereby eliminating the risk of movement jumps caused by kinematic limits or target drift upon resuming control.  Overall it acts as an "anchor reset" operation that ensures every manipulation session remains intuitive and easy to operate. This feature is also useful when manipulating the robotic arm for tasks that require fine movement adjustments. For example, the operator can do a "fine-tune - motion pause - continue fine-tune" cycle behavior. This functionality could also potentially be implemented as a "dead man's switch", but doing so makes simultaneous gripper control operation more difficult, thus we retain the current choice.

\section{Usage Note}
In addition to the introduced control logic and basic functionalities, Quest2ROS2 also possesses several new code-related features and alternative usage methods that make development and operation based on our framework more convenient, which will be discussed in this section. Note that information about the installation and basic usage of the framework will not be covered, as they are detailed in the code's documentation.

\subsection{Features}

\textbf{Core Architecture and Extensibility.} Most of the core functionality of our framework is integrated into the \textit{BaseArmController} class, which includes dedicated methods for configuring the necessary topics for ROS 2 node operation. The implementation of the control nodes for the left and right arms in our dual-arm teleoperation system is achieved simply by inheriting this base class and modifying the relevant topic interfaces. Therefore, if a user needs to adjust the framework for a different robot setup, the simplest way is to directly change the relevant topics in the base class or inherit this class and override the parameter setting methods.

\textbf{Debugging Tools.} Similar to Quest2ROS, our framework uses ROS-TCP-ENDPOINT\footnote{https://github.com/Unity-Technologies/ROS-TCP-Endpoint/tree/main-ros2} packets to establish communication between the device and the ROS 2 system. On top of that, to ensure robustness and predictable testing during development, we created a dedicated script, \textit{SimulationInput}, that allows for the external publication of the Quest hand pose topics \textit{PoseStamped}, effectively bypassing the physical VR controller. This capability is used to send predefined, clean movement trajectories to the system, enabling precise observation of the robot's reaction to ideal inputs and eliminating the noisy data and human jitter inherent to live teleoperation, thereby simplifying the debugging of the control pipeline. We have also added nodes that detect the controller connection status in real-time to facilitate user inspection, which can be found in \textit{CheckTCPconnection} node.

\begin{figure}[t]
  \centering
  \includegraphics[width=\linewidth]{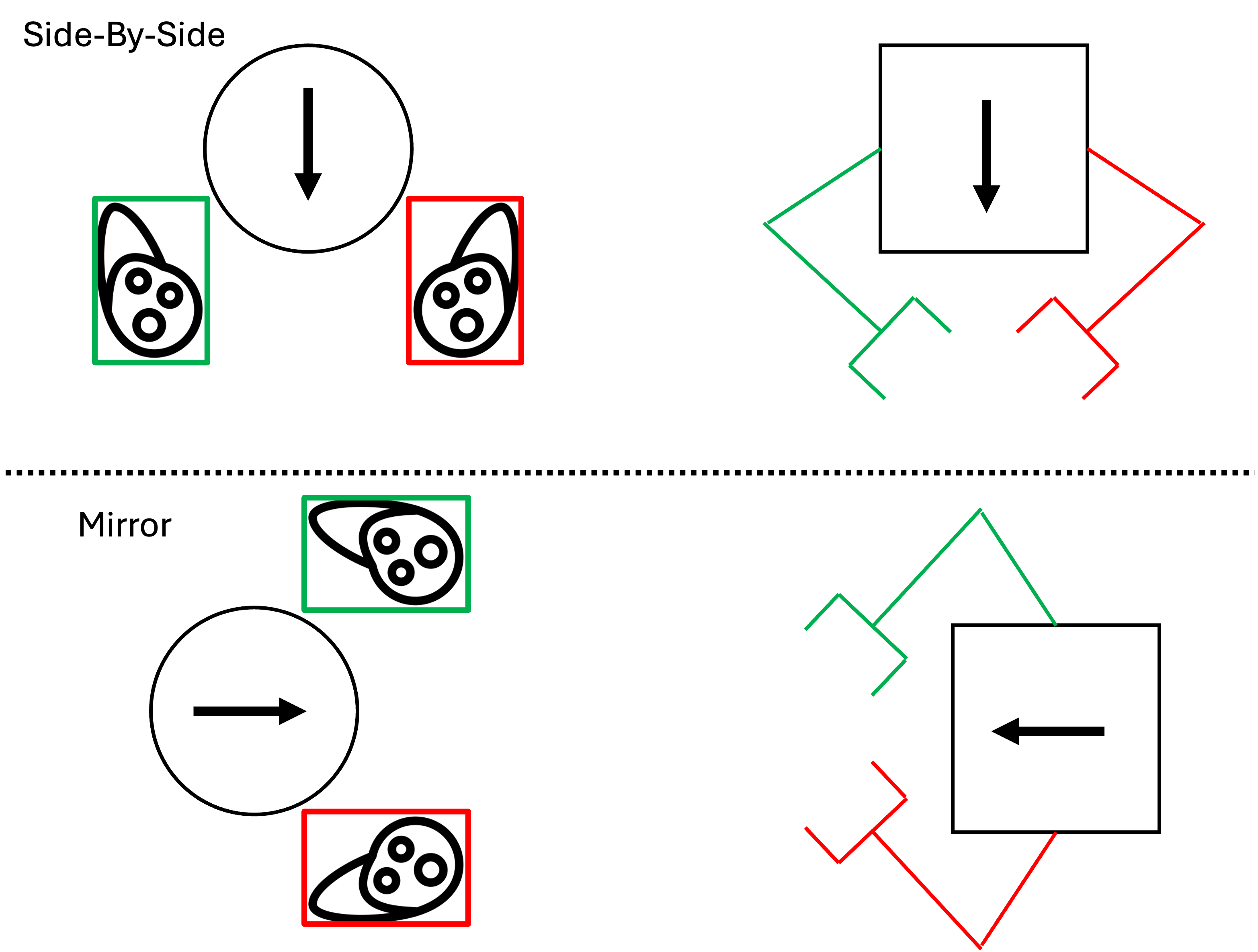}
  \caption{Demonstration of Control Methods Affected by Orientation and Position. The circle represents the operator, the square represents the robot, and the arrows indicate their respective orientations (facing directions). The colors on the controllers and robotic arms identify the one-to-one correspondence between them. In the "Side-by-Side" stance, the chirality (handedness) of the operator and the robotic arms is consistent, whereas in the "Mirror" scenario, the chirality between the controller and the robotic arm is symmetrical.}
  \label{fig:mirror}
\end{figure}

\textbf{"Side-by-Side" Control and "Mirror" Control}: For the teleoperation of bi-manual robots, operators sometimes need to handle the issue of chirality between the controllers and the robot end-effectors, based on the scenario. Specifically, the simplest dual-arm teleoperation setup, which we refer to as the "Side-by-Side" method, as shown in upper part of Figure \ref{fig:mirror}, involves the operator standing beside the robot, facing the same direction, and ensuring that the left and right controllers correspond to the left and right end-effectors, respectively, to maintain operational consistency. While this method is similar to standard single-arm teleoperation and intuitive, it significantly restricts the operator's line of sight. For certain tasks requiring fine manipulation, the operator's view of the precise operation may be obstructed by the robot arms. This phenomenon is particularly pronounced when using large-sized robots such as the Kuka iiwa7.

While testing our new control logic, we found that a more user-friendly mode of operation is the "Mirror" stance, as illustrated in the bottom half of Figure \ref{fig:mirror}. Given the structural design of the framework, this change can be achieved simply by swapping the interface topics of the left and right arm controller nodes.

\subsection{Safety Guidelines}
Given the inherent safety risks in robot teleoperation, we recommend correct configuration and maintaining constant safety awareness when using the framework. We have the following two suggestions to avoid introducing additional safety issues.

Due to the inherent characteristics of the Meta Quest devices \cite{welle2024Quest2ROS}, it is recommended that users keep both controllers visible to the headset at all times to maintain tracking accuracy. However, through testing in a dual-arm environment—especially in the "Mirror" operation mode previously discussed—finding a suitable position to place the headset can sometimes be slightly challenging. We recommend that if users struggle to find such a suitable position, the safest method is to wear the headset angled upwards on the top of the user's head. This is primarily to ensure that the device's accuracy does not degrade during teleoperation, which could cause the controller to send random commands and lead to unexpected behavior from the robotic arm.

Furthermore, with the introduction of functionalities for the up and down buttons on the controllers. We found that when a user presses both buttons simultaneously during the teleoperation node's runtime (intending to use Quest2ROS's native coordinate alignment feature), the system sometimes struggles to prioritize between our custom button functions and the coordinate alignment function. For safety, we recommend users perform the coordinate alignment for the left and right controllers only once after setting up the Quest2ROS app and before starting the ROS2 nodes, and do not repeat this operation during subsequent runtime.

\section{Future Work}
The current framework also provides several feasible avenues for future development to enhance its capabilities in dual-arm VR teleoperation and data collection:
\begin{itemize}
    \item One critical path is to implement a more advanced gripper control logic. Currently, the framework supports a simple open/close toggle. Extending this to offer flexible control over the gripper's finger position would enable the teleoperation system to be applied to more intricate tasks requiring specific grasp widths, such as manipulating thin or delicate objects.
    \item Another important direction is the integration of secondary data collection functions. This includes core features for robot learning, such as easily starting and stopping data recording using a dedicated button command. Additionally, implementing a robot homing (return-to-initial-pose) function would significantly improve safety and workflow efficiency, allowing operators to quickly reset the system between trials or in case of an error state. 
    \item A more robust controller button priority logic can also be implemented for a safer teleoperation experience.
\end{itemize}

\begin{acks}
This research is funded by the Wallenberg AI, Autonomous Systems and Software Program (WASP). We thank Faseeh Ahmad and Marcus Klang for their continued help and discussion during development.
\end{acks}

\bibliographystyle{ACM-Reference-Format}
\bibliography{references}


\end{document}